\newif\ifarxiv%
\newif\ifralfinal%
\arxivfalse%
\ralfinaltrue%
\ifralfinal%
    \documentclass[letterpaper, 10 pt, journal, twoside]{IEEEtran}%
\else%
    \documentclass[letterpaper, 10 pt, conference]{ieeeconf}%
\fi%
\IEEEoverridecommandlockouts%
\pdfoutput=1%
\usepackage{amsmath,amssymb,amsfonts}%
\usepackage{algorithmic}%
\usepackage{algorithm}%
\usepackage{array}%
\usepackage{textcomp}%
\usepackage{stfloats}%
\usepackage{verbatim}%
\usepackage{graphicx}%
\graphicspath{{../figures/}}%
\DeclareGraphicsExtensions{.pdf,.jpeg,.png}%
\usepackage{cite}%
\usepackage{booktabs}%
\usepackage{hyperref}%
\usepackage[table,xcdraw]{xcolor}%
\usepackage{multirow}%
\usepackage{array}
\usepackage{tcolorbox}
\usepackage[super]{nth}%
\usepackage[top=60pt,bottom=50pt,left=50pt,right=50pt]{geometry}
\hypersetup{
  colorlinks=true,
  linkcolor=blue!80!black,
  urlcolor=blue,
  citecolor=green!50!black,
  linktocpage
}
\usepackage{caption}
\newcommand{\ocMCSMRT}[4]{\multicolumn{#1}{#2}{\multirow{#3}{12mm}{\centering{\B{#4}}}}}

\newcommand{\ocblbT}{\multicolumn{2}{|l|}{}}

\newcolumntype{M}[1]{>{\centering\arraybackslash}m{#1}}
\newcolumntype{L}[1]{>{\raggedright\arraybackslash}m{#1}}

\newcommand*{\B}[1]{\ifmmode\boldmath{#1}\else\textbf{#1}\fi}
\newcommand*{\I}[1]{\ifmmode\mathit{#1}\else\textit{#1}\fi}

 \DeclareMathOperator{\ocfor}{for}

\newcommand{\mdots}{\hbox to 1em{.\hss.\hss.\hss}}
\def\expOneAggMCVPR {\mbox{$\approx\!41\%$}}
\def\expOneAggMCSVM {\mbox{$\approx\!46\%$}}
\def\expOneAggMCNN {\mbox{$\approx\!55\%$}}
\def\expOneAggMeanVPR {\mbox{$\approx\!9.8m$}}

\def\expOneAggMeanNN {\mbox{$\approx\!3.1m$}}

\def\expTwoAggPrecVPR {\mbox{$\approx\!97\%$}}

\def\expTwoAggPrecNN {\mbox{$\approx\!99\%$}}
\def\expTwoAggMeanVPR {\mbox{$\approx\!2.0m$}}

\def\expTwoAggMeanNN {\mbox{$\approx\!0.5m$}}

\def\numLayers {$4$}
\def\numLayerNeurons {$128$}
\def\batchSize {$8$}
\def\learnRate {$0.00001$}
\def\dropoutRate {$10\%$}

\def\ootAPGEM   {$=43.6\%$}
\def\ootNETVLAD {$=47.2\%$}
\def\ootSALAD   {$=13.3\%$}

\def\alphaAPGEM   {$\alpha=6$}
\def\alphaNETVLAD {$\alpha=3$}
\def\alphaSALAD   {$\alpha=35$}

\def\precAPGEM   {$91.6\%$}
\def\precNETVLAD {$88.2\%$}
\def\precSALAD   {$98.6\%$}

\def\rcllAPGEM   {$21.3\%$}
\def\rcllNETVLAD {$29.3\%$}
\def\rcllSALAD   {$63.6\%$}

\ifarxiv%
    \usepackage{fancyhdr}%
\fi%

\begin{document}
    \title{\ifarxiv\LARGE\bf\fi
    Improving Visual Place Recognition Based Robot Navigation By Verifying Localization Estimates
    }%
    \author{%
        \parbox{\linewidth}{\centering
            Owen~Claxton$^{1}$, %
            Connor~Malone$^{1}$, %
            Helen~Carson$^{1}$, %
            Jason~J.~Ford$^{1}$, %
            Gabe~Bolton$^{2}$, %
            Iman~Shames$^{2}$, %
            Michael~Milford$^{1}$~\IEEEmembership{Senior Member,~IEEE}%
        }%
        \ifralfinal%
            \thanks{Manuscript received: June 24, 2024; Revised: Sept 10, 2024; Accepted: Sept 27, 2024.}
            \thanks{This paper was recommended for publication by Editor Sven Behnke upon evaluation of the Associate Editor and Reviewers' comments.}%
        \fi%
        \thanks{This research is partially supported by an ARC Laureate Fellowship FL210100156 to M.Milford, the QUT Centre for Robotics, the Centre for Advanced Defence Research in Robotics and Autonomous Systems, and received funding from the Australian Government via grant AUSMURIB000001 associated with ONR MURI grant N00014-19-1-2571. The work of C.Malone and H.Carson was supported in part by an Australian Postgraduate Award.}%
        \thanks{$^{1}$O.Claxton, C.Malone, H.Carson, J.Ford, and M.Milford are with the QUT Centre for Robotics, School of Electrical Engineering and Robotics at the Queensland University of Technology, Brisbane, Australia (e-mail: \{o.claxton, cj.malone, h.carson, j2.ford, michael.milford\}@qut.edu.au).}%
        \thanks{$^{2}$G.Bolton and I.Shames are with the CIICADA Lab, College of Engineering, Computing and Cybernetics, at the Australian National University, Canberra, Australia (e-mail: \{gabe.bolton, iman.shames\}@anu.edu.au).}
        \ifralfinal%
            \thanks{Digital Object Identifier (DOI): see top of this page.}%
        \fi%
    }

    \ifralfinal
        \markboth{IEEE Robotics and Automation Letters. Preprint version. Accepted October, 2024. DOI: \href{https://doi.org/10.1109/LRA.2024.3483045}{10.1109/LRA.2024.3483045}}%
        {Claxton \MakeLowercase{\textit{et al.}}: Improving Visual Place Recognition Based Robot Navigation By Verifying Localization Estimates}
    \fi
    
    \maketitle%
    
    \ifarxiv
        \thispagestyle{fancy}
        \pagestyle{plain}
    \fi
    \begin{abstract}
        Visual Place Recognition (VPR) systems often have imperfect performance, affecting the `integrity' of position estimates and subsequent robot navigation decisions. Previously, SVM classifiers have been used to monitor VPR integrity. This research introduces a novel Multi-Layer Perceptron (MLP) integrity monitor which demonstrates improved performance and generalizability, removing per-environment training and reducing manual tuning requirements. We test our proposed system in extensive real-world experiments, presenting two real-time integrity-based VPR verification methods: a single-query rejection method for robot navigation to a goal zone (Experiment 1); and a history-of-queries method that takes a best, verified, match from its recent trajectory and uses an odometer to extrapolate a current position estimate (Experiment 2). Noteworthy results for Experiment 1 include a decrease in aggregate mean along-track goal error from \expOneAggMeanVPR~to \expOneAggMeanNN~, and an increase in the aggregate rate of successful mission completion from \expOneAggMCVPR~to \expOneAggMCNN. Experiment 2 showed a decrease in aggregate mean along-track localization error from \expTwoAggMeanVPR~to \expTwoAggMeanNN, and an increase in the aggregate localization precision from \expTwoAggPrecVPR~to \expTwoAggPrecNN. Overall, our results demonstrate the practical usefulness of a VPR integrity monitor in real-world robotics to improve VPR localization and consequent navigation performance.

    \end{abstract}
    
    \ifralfinal
    \fi

    \section{Introduction}\label{sec:intro}

    \ifralfinal
    \IEEEPARstart{I}{n}
    \else
    In
    \fi
    many deployment scenarios, utilizing a pre-generated map is an effective enabler for position estimation (also referred to as localization). Once created, a map can be shared with any robots operating within the surveyed environment. This can be advantageous from a performance perspective and can yield a simpler implementation with reduced requirements and problem complexity, however, it does create additional overheads for map generation. Position estimation using pre-generated maps is useful in cases where robots may retrace routes through an environment repeatedly, such as warehouse operations or underground mines \cite{Jacobson21}.\par
    Visual Place Recognition (VPR) is a well-established method using pre-generated maps, whereby a best-matching image can be retrieved from a database (or map) to yield a position estimate using associated geographical metadata \cite{Lowry2016,Schubert23_VPR_Tutorial,Masone2021}. However, translation and orientation errors from incorrect matches can lead to catastrophic errors in navigation decision-making. Consequently, to enable VPR deployment in active navigation scenarios, there is a need for trustworthy verification systems which can prevent unsafe system behavior. In this letter, we present research addressing this problem and integrating it into a robot navigation system (Figure \ref{fig:MainDiagram}).
    
    In particular, we make the following contributions:
    \begin{figure}[!t]
        \centering
        \includegraphics[width=\linewidth]{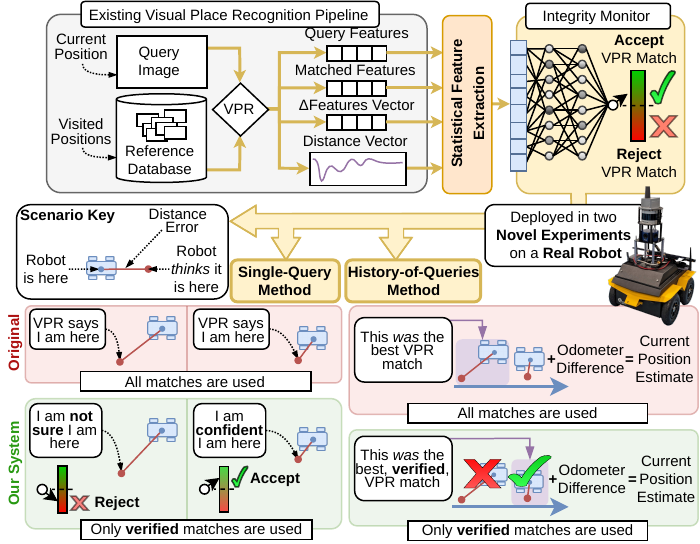}
        \caption{Overview of our system, demonstrating how our addition of a predictive verification system (a Multi-Layer Perceptron (MLP) network) results in safer navigation. See Figure~\ref{fig:MLPschem} for more.}
        \label{fig:MainDiagram}
        \vspace*{-\baselineskip}
    \end{figure}
    \begin{enumerate}
        \item We close the loop on an integrity monitor for verification previously only demonstrated passively on datasets, integrating it into an active robot navigation system to make navigational decisions.
        \item We present a novel approach using a Multi-Layer Perceptron network for VPR integrity monitoring, rather than the previously adopted Support Vector Machine (SVM), improving performance, generalizability across datasets, and outcomes in robot navigation paradigms.
        \item We present a new, more capable method that applies verification processes to recent historical location estimates, then projects forwards from the best, verified match to estimate a current location using odometry.
        \item We evaluate the performance of the proposed techniques in extensive mobile robot navigation experiments across a variety of indoor and outdoor environments and conditions, and propose two new robot navigation paradigms where performance is dependent on localization.
        \item We apply the proposed techniques on a variety of well-established and state-of-the-art VPR techniques.
    \end{enumerate}
    We define verification as a process for assessing whether the VPR system output should be retained or rejected.
    
    The letter proceeds as follows: in Section \ref{sec:background} we review the literature, in \ref{sec:approach} we detail our approach, in \ref{sec:experiments} we introduce our experimental methods, and in \ref{sec:results} we discuss the results.
    
    Our code and dataset features are publicly available\footnote{\label{githubrepo}\url{https://github.com/QVPR/aarapsiproject}}.
    
    \section{Background}\label{sec:background}
    Here, we provide a brief overview of VPR and introduce the concept of localization integrity as it relates to our work.

    \subsection{Visual Place Recognition}
        Typically VPR involves extracting features from a current (query) image and comparing these with features from a set of reference images taken along a previously-traversed path to determine a best-match\cite{Schubert23_VPR_Tutorial}. Numerous VPR techniques exist, with feature extraction and matching varying in complexity and performance \cite{NetVLAD18, Mereu2022, Wang22_CVPR, PatchNetVLAD, CosPlace, EigenPlaces23, Lsplacerecgss},  noting the recent techniques MixVPR \cite{MixVPR23}, DinoV2 SALAD \cite{optimalsalad}, and AnyLoc \cite{anyloc}.

        When deployed on real-world navigating robots, VPR is susceptible to potentially large localization errors due to varied environmental conditions, lighting, and presence of dynamic objects \cite{2021Schubert}. Although stand-alone systems have no match validation, some research has attempted to characterize the uncertainty. Uncertainty is often modelled using known distributions~\cite{zaffar2024estimation}, however, other work shows VPR errors are often discontinuous and non-Gaussian \cite{Zhu22}. Complementary to this, \cite{trivigno2023divide} had some success correlating VPR accuracy (within 25m) with network outputs by re-framing VPR as a classification problem. Sequence matching \cite{Stenborg20, Yin19,Tsintotas2021}, Bayesian techniques \cite{Xu21, Warburg21, Warburg23} and particle filters \cite{Li15,particlefilters18} may be used to help reduce error but adds cost in complexity and sensing. Some research has focused on predicting which VPR combinations will perform best in a given environment \cite{hausler2021unsupervised,Malone23}, but these methods are computationally intensive.

    \subsection{Localization Integrity}
        Localization integrity refers to the level of trust we can place in a system to produce a localization estimate within an acceptable predefined error tolerance \cite{Hage21}. Integrity monitoring is well-established for GNSS navigation systems, but has received little attention for visual navigation systems \cite{Zhu22}. 
    
        Some integrity monitoring approaches have been proposed for stereo visual odometry \cite{Li2019, Wang20, Fu2020, Fu22}; or for fusing visual odometry with sensors such as GNSS, LiDAR and radar \cite{Gabela20, Calhoun16, Arana2020, Balakrishnan2020}. Other approaches include vision-based lane detectors \cite{AlHage19}, landmarks localization \cite{Bhamidipati21}, or objects of interest localization within a scene \cite{Calhoun16}. However, none of these methods specifically address VPR. In our previous work \cite{Carson22}, we proposed SVM-based integrity prediction methods for VPR such that each match's accuracy is predicted to be within a tolerance; but this was investigated passively on datasets, not actively navigating robots. Regardless of method, integrating an integrity monitor is attractive as it can enable a robot to introspectively improve real-time operational decisions.
    \section{Approach}\label{sec:approach}
In this section we describe the proposed single-query and history-of-queries (HoQ) localization methods.
    
    \subsection{VPR Technique Requirements}\label{sec:approach_VPR_tech}
        The integrity monitor we build on from~\cite{Carson22} only requires some form of distance vector input for predictions, making it compatible with many, if not all, VPR techniques. The distance vectors consist of mathematical distances between query image features and those from a reference set; we refer to these distances as `match distances'. In this work, we maintain compatibility with all, or at least the vast majority of, VPR techniques by similarly using information already available from the VPR process to verify VPR integrity. We demonstrate this by testing using three VPR techniques: AP-GeM~\cite{Revaud2019APGEM}, NetVLAD~\cite{NetVLAD18}, and SALAD~\cite{optimalsalad}.
    
    \subsection{Integrity Prediction}\label{subsec:integrity_prediction}
        For any given VPR technique, we predict the integrity of VPR matches using a similar supervised learning process to \cite{Carson22}. That is, we generate VPR matches for a query and reference traverse separate from the test route to use as training data. Then, we use ground truth correspondences to train a predictor to classify the training VPR matches as in-tolerance or out-of-tolerance based on features extracted from the distance vectors and VPR features for each query. Importantly, in this work, we introduce the first use of a neural network for the VPR integrity monitoring task; this increases generalizability by removing per-environment training requirements and reducing manual tuning.
        
\begin{figure}[!ht]
            \centering
            \includegraphics[width=0.9\linewidth]{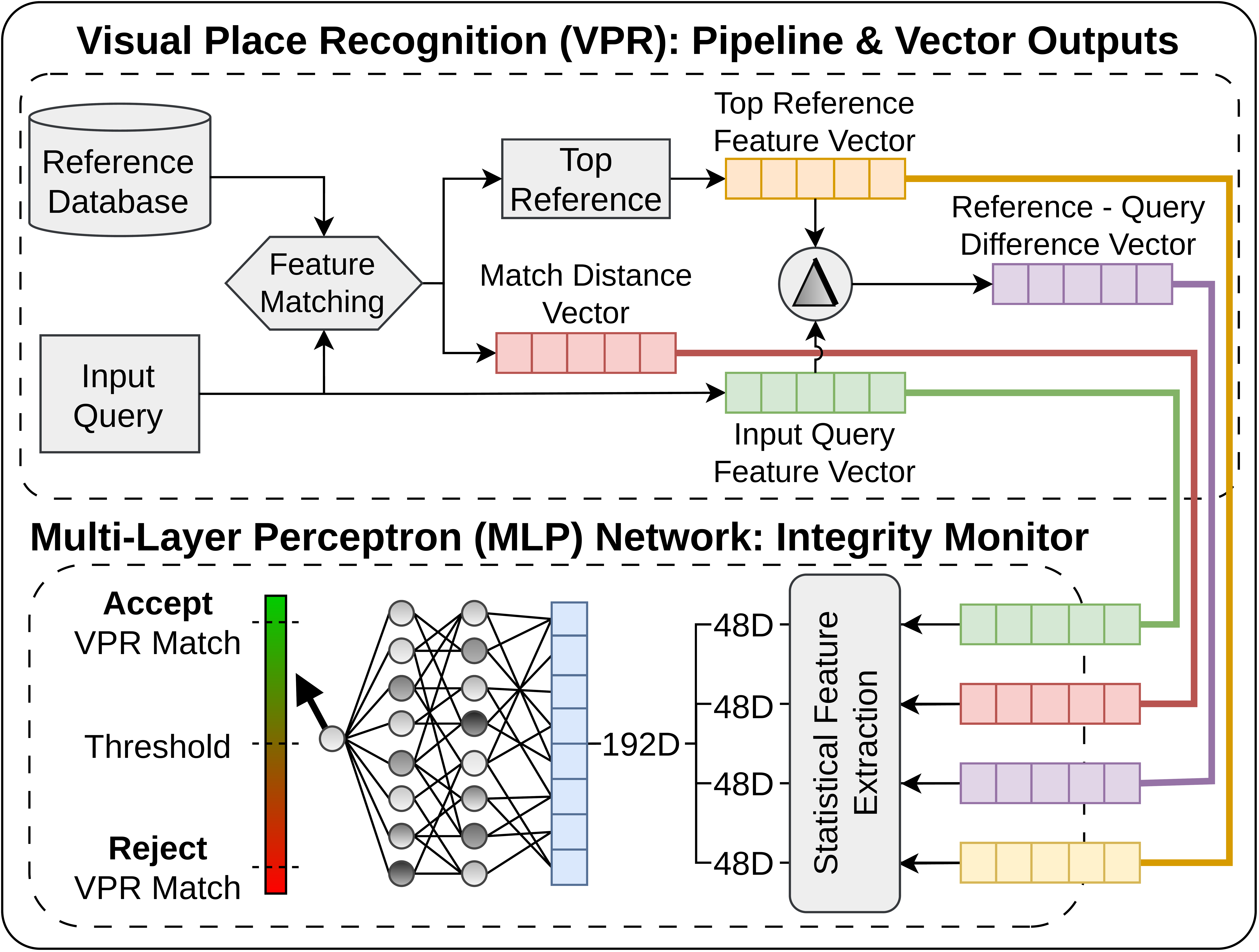}
            \caption{An overview of how inputs for the MLP integrity monitor are extracted. For a given query, we use four vectors formed from the output of the VPR process. The vectors pass through a statistical feature extractor and are then concatenated. These steps guarantee the size of the input is suitable for the MLP network, which accepts a 192-dimensional vector.}
            \label{fig:MLPschem}
            \vspace*{-\baselineskip}
        \end{figure}

        We improve on the previously adopted SVM model from \cite{Carson22} by implementing a small multi-layer perceptron (MLP) network, with a simple fully-connected structure, to utilise the increased learning capability of neural networks whilst maintaining the light-weight nature of the predictor. We use 48 hand-crafted statistical feature calculations which, when applied to four vectors, yields a total of 192 unique features as the classification input. For brevity we include a simplified diagram of the process (Fig.~\ref{fig:MLPschem}); the detailed mathematical formulation and the software implementation is provided in the mentioned repository\textsuperscript{\ref{githubrepo}}. We use such a large number of statistical features to remove the necessity of knowing the best features for individual VPR techniques.

    \subsection{MLP Integrity Monitor Training}
    \renewcommand{\thefootnote}{\fnsymbol{footnote}}
    For a given query image with index $k$, the input vector for the MLP network becomes a 192-dimensional vector created through concatenating four groups of 48 statistical features. These groups are calculated from the $k^{\text{th}}$: distance vector, $\mathbf{D}$\footnote[7]{Bold denotes arrays. Subscripts denote indices, which start from $1$.}, as per \cite{Carson22}; the query VPR feature vector, $\mathbf{Q}$; the best-match reference feature vector, $\mathbf{R}$; and the VPR feature difference vector, $\mathbf{V}=\mathbf{R}-\mathbf{Q}$. The query training label, $\mathbf{P}_k$, simply becomes a binary value indicating whether the top VPR match corresponds to a reference in-tolerance, $1$, or out-of-tolerance, $0$. We formulate the problem as a regression task using a weighted mean-squared-error (MSE) loss, providing control over the rate queries are predicted as out-of-tolerance. The loss function, $L(\mathbf{P},\mathbf{\hat{P}},\alpha)$, is defined as:
    \renewcommand{\thefootnote}{\arabic{footnote}}
    \begin{equation}
            L(\mathbf{P},\mathbf{\hat{P}},\alpha) = \frac{1}{N}\cdot\sum_{k=1}^{N}
            \begin{cases}
                (\mathbf{P}_k-\mathbf{\hat{P}}_k)^2 & \mathbf{P}_k = 1 \\
                \alpha(\mathbf{P}_k-\mathbf{\hat{P}}_k)^2 & \mathbf{P}_k = 0
            \end{cases} \ ,
    \end{equation}
    where $\mathbf{\hat{P}}_k$ is the integrity prediction from the MLP network, $\alpha$ is a scalar weight, and $N$ is the dimension of $\mathbf{P}$ and $\mathbf{\hat{P}}$. We set $\alpha > 1$ to produce a cautious integrity monitor that reduces the number of out-of-tolerance VPR matches that are incorrectly predicted as in-tolerance (false positives), given these pose a greater risk to robot navigation tasks. Effectively, $\alpha$ controls the precision and recall of classification performance for the integrity prediction task.
    
    \subsection{Localizing with VPR and Integrity: Single-Query Method}\label{subsec:localizing}
        In the single-query scenario where each query is treated independently, the VPR system performs query-to-reference matching and provides a position estimate using the retrieved 2D pose ($x$,$y$,$\theta$) from where the best-match reference image was collected. The integrity monitor (Section \ref{subsec:integrity_prediction}) then verifies the unprocessed output by only accepting position estimates that are predicted to be in-tolerance of the true position; all predicted out-of-tolerance estimates are rejected.

        \begin{figure}[!ht]
            \centering
            \includegraphics[width=\linewidth]{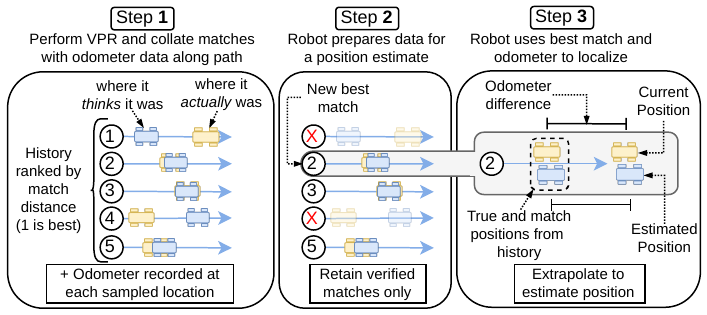}
            \caption{HoQ method: we rank a recent history of VPR matches by their match distance (1), reduce to verified matches (2), then extrapolate from the best, verified match to estimated position using the odometer difference (3).}
            \label{fig:exp2diagram}
            \vspace*{-\baselineskip}
        \end{figure}%

    \subsection{Utilizing Historical Data: History-of-Queries Method}\label{subsec:util_hist_data}
        We can improve the single-query localization decisions (Section \ref{subsec:localizing}) by utilizing an odometer to extrapolate a position estimate based on a recent history of predicted in-tolerance VPR matches, which we refer to as the history-of-queries (HoQ) method. The use of historical information is a well-known component of many navigation systems: our key contribution here is the added precursor step where all entries in the history are verified (such that only predicted in-tolerance matches are retained), distinguishing our method from those which only perform statistical or threshold filtering. If the system finds at least one verified historical match, then it proceeds to localize. We diagram this in Figure \ref{fig:exp2diagram}.

        This process can be described by the following. Consider some unknown current vehicle pose $\chi$ and known odometer scalars $\omega$. Within the previous $d$ meters, we sample VPR match indices $\mathbf{M}$ and odometer scalars $\mathbf{W}$, collected together at irregular spatiotemporal intervals. For each match index $\mathbf{M}_i$, we find a corresponding: match distance $\mathbf{D}_i$, integrity prediction $\mathbf{\hat{P}}_i$, and pose estimate $\mathbf{X}_{\mathbf{M}_i} \in \mathbf{X}$, where $\mathbf{X}$ is the ordered set of 2D poses with elements indexed to each reference image (dimension $n$). We assume $\chi \equiv \mathbf{X}_{q}$, where $q$ is the index of the closest (Euclidean) pose in $\mathbf{X}$ to $\chi$. Using $\mathbf{M}, \mathbf{D}, \mathbf{W}, \mathbf{\hat{P}}$, and $\mathbf{X}_{\mathbf{M}}$ as the recent history contents, per Figure~\ref{fig:exp2diagram}, we now work to determine $q$.
        
        We denote the best match index as $\mathbf{M}_b$, i.e., the historical match with the lowest match distance (as per Figure~\ref{fig:exp2diagram}, Step 2); $\Delta$ denotes the distance travelled since $\mathbf{M}_b$ was recorded:
        \vspace{-0.5\baselineskip}
        \begin{align}
            b &= \arg\min(\mathbf{D}) \label{eq:e2-B} \\ 
            \Delta &= | \omega - \mathbf{W}_b | \ .
        \end{align}
        To determine $q$, we first need the along-track distances between each reference image pose\footnote{For brevity, we ignore additional complexities such as traverse loops.} starting from the best match index $b$, denoted by $\mathbf{T}$:
        \vspace{-0.5\baselineskip}
        \begin{align}
            \begin{split}
                \mathbf{T} =&~(0, \left\lvert \mathbf{X}_{b+1} - \mathbf{X}_{b}\right\rvert, |\mathbf{X}_{b+2} - \mathbf{X}_{b+1}|, \\
                &\quad\mdots,|\mathbf{X}_{n} - \mathbf{X}_{n-1}|) \ .
            \end{split}
        \end{align}
        We then compute the running sum of the elements of $\mathbf{T}$, denoted as $\mathbf{S}$, from which we find $q$ and $\chi$ per Fig.~\ref{fig:exp2diagram}, Step 3:
        \begin{align}
            \mathbf{S}_j &= \sum_{i=1}^{j} \mathbf{T}_{i}\hspace{2mm}\ocfor j \in \{1,...,n-b+1\} \ , \\
            q &= b + \left(\arg\min \left\lvert \Delta - \mathbf{S} \right\rvert\right) - 1 \ .
        \end{align}
        To implement verification, we replace $\mathbf{D}$ from \eqref{eq:e2-B} with $\mathbf{\hat{D}}$ which uses $\mathbf{\hat{P}}$ to remove predicted out-of-tolerance matches:
        \begin{align}
            \mathbf{\hat{D}}_i &=
            \begin{cases}
                 \mathbf{D}_i & \mathbf{\hat{P}}_i = 1 \\
                 1 + \max \mathbf{D} & \mathbf{\hat{P}}_i = 0
            \end{cases} \label{eq:e2-Dp} \\
            \hat{b} &= \arg\min(\mathbf{\hat{D}}) \ .
        \end{align}
        If $\mathbf{\hat{D}}_{\hat{b}} > \max \mathbf{D}$, then $q$ cannot be determined, and so the system declines to provide a localization estimate.
    \section{Experimental Procedure}\label{sec:experiments}
    In this section, we introduce our two experiment designs and items such as hardware, datasets, parameters, and metrics.

    \vspace{-\baselineskip}
        
    \subsection{Experiment 1 Design}\label{subsec:exp1design}
        
\begin{figure}[!ht]
            \centering
            \includegraphics[width=0.9\linewidth]{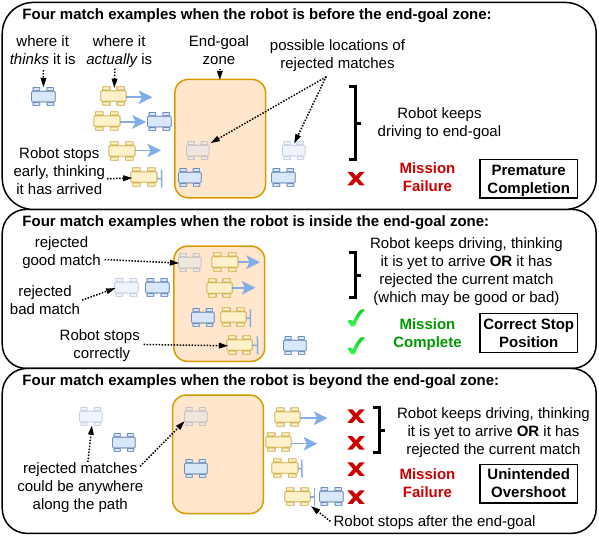}
            \caption{In Experiment 1, the robot moves towards the end-goal, and makes a navigation decision for each verified VPR match.}
            \label{fig:exp1diagram}
        \end{figure}

        In the first experiment scenario, a robot moves along a path whilst attempting to autonomously determine if it has reached an end-goal zone. We implement the single-query method introduced in Section \ref{subsec:localizing} for localization. The mission is successfully completed if the robot correctly identifies it has arrived at the end-goal within an assessment tolerance of $\pm$0.5m, otherwise the mission is failed, as shown in Fig. \ref{fig:exp1diagram}.
        
        For fair assessment, given the underlying baseline system performance changes along the route due to environmental and dynamic variations, we randomly select 50 start locations over the reference path, with end-goal locations allocated at 5m, 10m, 25m and 50m from each starting position. As navigation decisions require a tighter tolerance, the robot will flag arrival if its localization estimate is anywhere beyond the point 10cm before the end-goal center. This provides a margin of error in achieving the looser, actual, target tolerance. We emphasize that VPR localizes from a reference set of the whole environment, not the subset being traversed.
        
        When our integrity monitor is used, only localization estimates along the route that are verified as in-tolerance will be acted upon, and all predicted out-of-tolerance localization estimates will be rejected. The practical implication of this is that if a robot can identify and reject all VPR failures along a route, then it will only act on in-tolerance localization estimates and should theoretically finish closer to the end-goal. However, in the event that the monitor is overly cautious and rejects many in-tolerance estimates, then the robot will overshoot the end-goal until the next verified estimate is found, thus failing the mission. In this way, we can assess the impact of both the precision and recall of our proposed prediction system from a practical perspective.
    
        \subsubsection{Performance metrics}\label{subsubsec:exp1_perf_metrics}
            We use absolute along-track error in meters between the end-goal and ground truth finish location (for missions where the robot reports it has found the goal) as the key performance metric, denoted as `goal error', along with percentage of successful mission completions.

    \vspace{-\baselineskip}
    \subsection{Experiment 2 Design}\label{subsec:exp2design}   

        In the second experiment scenario, a robot repeats a pre-mapped traverse whilst localizing using the HoQ method from Section \ref{subsec:util_hist_data}. In this scenario, the goal is to maximize precision. We select a history length of 1.5m. For fair analysis the robot must initialize the history by traversing this distance.  
        
        \subsubsection{Performance Metrics}\label{subsubsec:exp2_perf_metrics}
            We use absolute along-track error in meters between the estimated position and ground truth position as the key performance metric, denoted as `localization error', along with system precision and recall.

    \vspace{-\baselineskip}
    \subsection{Platform Overview}\label{subsec:plat_overview}
        For this work, we used a Clearpath Robotics Jackal platform. Whilst our approach can function with a selection of sensor modalities, our particular implementation used a monocular camera feed and odometer, which we extracted from the forward vision of an Occam 360 panoramic camera (resulting in an RGB feed, dimensions 720 by 480 pixels  and a horizontal field-of-view of 72 degrees) and wheel encoder data respectively. Using data from a Velodyne VLP-16 LiDAR, we employed HDL Graph Slam \cite{hdl_packages} to generate a 3D pointcloud map of each environment offline, and HDL localization \cite{hdl_packages} to localize the robot within the map during testing and data collection. This provided the necessary accuracy for quantitative analysis, acting as a ground truth equivalent. The platform, including sensors, were additionally simulated in the Robot Operating System (ROS) Gazebo environment.

    \vspace{-\baselineskip}    
    \subsection{Data Collection: Environments and Conditions}\label{subsec:data_col_var}
\begin{figure}[!ht]
            \vspace*{-0.5\baselineskip}
             \centering
            \includegraphics[width=0.9\linewidth]{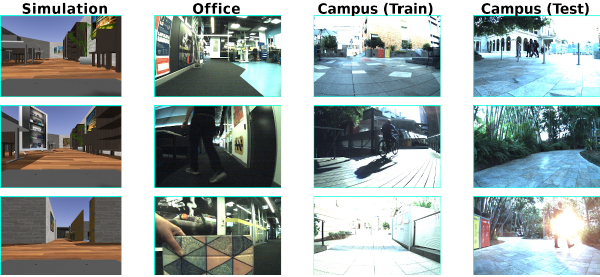}
            \caption{Photo bank of sample dataset images, grouped into columns per environment. Adversities include anomalous obstacles, dynamic objects, lighting changes, camera occlusion, glare, and time-of-day changes.}
            \label{fig:jackal}
        \end{figure}

        We conducted both experiments in two different environments: Office, containing indoor images from the QUT Centre for Robotics (QCR); and Campus, containing outdoor images from the QUT Gardens Point campus. For both environments we recorded two sets of conditions: Normal, where there is a typical presence of adversity given the environment (some minor lighting variation and dynamic or static objects); and Adverse, where there is a heightened presence of adversity, such as increased lighting variation (morning to evening), and more dynamic or static objects. Data capture periods were selected to control these introduced adversities. We emphasize here that we train a \textit{single} MLP integrity monitor to use across \textit{all} environments and conditions.
        This is achieved through a representative training dataset of each environment, formed from two query and reference pairs, captured under Adverse conditions. For these we use simulation data for Office, and a section of the campus spatially separated from all test data for Campus. Across all our collected data, we include 13 out of the 23 total `difficult' VPR instances put forward by \cite{2021Schubert}; we provide examples in Figure \ref{fig:jackal}.

        For simplicity in analysing performance in this novel setting, lateral path error was limited in traverses by retracing the reference route using ground truth. In real scenarios, VPR must compensate for lateral shift, which will likely reduce the precision of integrity monitors. This could be somewhat mitigated by controlling training data to include representative samples with viewpoint shifts, however, we believe the additional information and statistical features in the proposed MLP approach should improve robustness over previous SVM approaches.

        During each traverse, we sampled images at 10 Hz with an average vehicle speed of 0.5m/s. In total, approximately 1500m over 12 traverses was recorded for these experiments, with individual route lengths of 60m, 70m, 210m, and 160m for the Office training, Office testing, Campus training, and Campus testing environments, respectively. We emphasize here that, separate to the previously discussed training requirements, only a \textit{single} reference traverse is required for a real deployment scenario, noting that the query `traverse' would be formed from the live sensor feed during deployment. 
        
    \subsection{Experimental Parameters}\label{subsec:parameters}
        We define out-of-tolerance error as greater than 0.5 meters for indoor environments and greater than 1.0 meters for outdoor environments, which represents the range of navigational precision required for prospective navigation tasks in these environments such as delivering goods to a location. The system was operated at 10 Hz.

    \subsection{Multi-Layer Perceptron Hyperparameters}
        Throughout the described experiments, we train a single MLP integrity monitor for each VPR technique using a shared structure. We set the number of layers to \numLayers, the number of neurons per layer to \numLayerNeurons, the output layer to a single neuron, with ReLU activations between layers and finish with a Sigmoid activation. For training purposes, we use a batch size of \batchSize, learning rate of \learnRate, and a dropout rate of \dropoutRate. Due to differences in the training label balance when using different VPR techniques (out-of-tolerance: AP-GeM \ootAPGEM, NetVLAD \ootNETVLAD, SALAD \ootSALAD), we set \alphaAPGEM, \alphaNETVLAD, \alphaSALAD, for these VPR techniques respectively.

    \vspace{-\baselineskip}
    \subsection{Comparisons}\label{subsec:comparisons}
        We compare the performance of our proposed system against baseline VPR with no verification, as well as the SVM integrity monitor from our previous work \cite{Carson22}. Additionally, we compare against naive thresholds which reject VPR matches above a specified match distance, as diagnostic indicators for the performance characteristics of our proposed system: $N_P$, with equal precision to our proposed system; and $N_R$, with equal recall to our proposed system. These thresholds are calculated during training for each VPR technique, and are consistent across all environments and conditions. They correspond to: \precAPGEM~ precision, \rcllAPGEM~ recall for AP-GeM; \precNETVLAD~ precision, \rcllNETVLAD~ recall for NetVLAD; and \precSALAD~ precision, \rcllSALAD~ recall for SALAD.
    \section{Experiment Results}\label{sec:results}

In this section, we present the results and discussion for the two experiments and some additional studies. We compare metrics as defined in Sections \ref{subsubsec:exp1_perf_metrics} and \ref{subsubsec:exp2_perf_metrics} between the baseline system (where all matches are retained), the naive thresholds (as diagnostic indicators), our previous SVM approach~\cite{Carson22} (trained per environment and per VPR technique), and our proposed system (trained only per VPR technique).
    
    \subsection{Experiment 1 Results} \label{para:exp1_results}
    
        Results for Experiment 1 show our proposed system yields two key performance improvements: an increase in the number of missions successfully completed, and a reduction in goal error for all missions where the robot reported it had arrived at the goal location.
                
        \begin{figure}[!ht]
            \centering
            \includegraphics[width=0.95\linewidth]{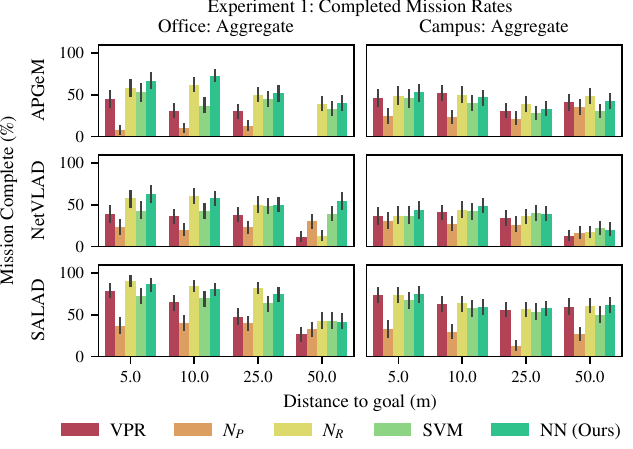}
            \caption{Missions completed for all techniques, aggregated by environment, for varying goal distances. A mission is completed when the robot stops within $\pm$0.5m of the end-goal.}
            \label{fig:exp1_key_results}
            \vspace*{-\baselineskip}
        \end{figure}

        Figure \ref{fig:exp1_key_results} shows the percentage of completed missions - where the robot successfully localizes within $\pm$0.5m of the end-goal - in aggregated Office and Campus, for traversal lengths between 5m and 50m. On average, across all experiments, our proposed system resulted in \expOneAggMCNN~ of missions completed, compared with \expOneAggMCVPR~ for the baseline system, and \expOneAggMCSVM~ for the SVM. Generally, longer traverses resulted in a lower number of mission completions.
        
        \begin{figure}[!ht]
            \centering
            \includegraphics[width=0.9\linewidth]{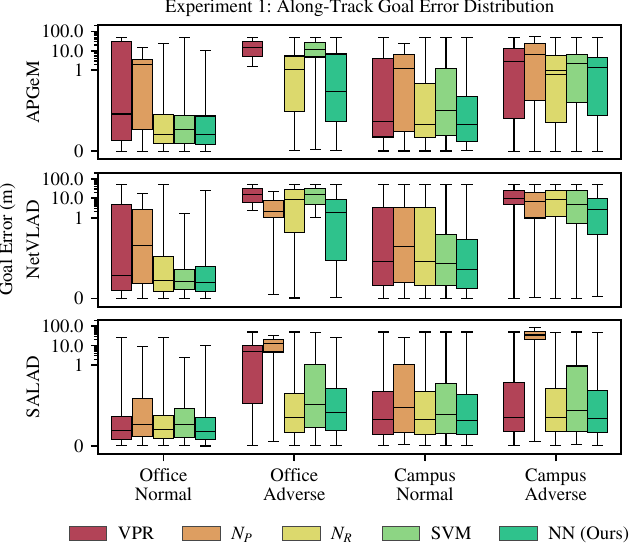}
            \caption{Goal error for all techniques, per environment and condition, aggregated for all end-goal distances (50 missions performed for each combination). For readability, a logarithmic scale applies for high errors.}
            \label{fig:exp1_loc_error}
        \end{figure}
        
        \begin{table}[!ht]
    \caption{Experiment 1 along-track goal error and percentage missions complete (M.C.). Best performance for each feature type is indicated by bold, ignoring naive thresholds. Naive thresholds are italicized if best performance.}
    \centering
    \scriptsize
    \setlength\tabcolsep{1.2mm}
    \begin{tabular}{lcL{11mm}|M{9mm}|M{9mm}M{9mm}|M{9mm}M{9mm}|}
        \cline{4-8}
        & & & \B{Baseline} & \multicolumn{4}{c|}{\B{Filtering Technique}} \\
        & & & \B{VPR}      & \boldmath{$N_P$} & \boldmath{$N_R$} & \B{SVM} & \B{Ours} \\
        \hline
        \ocMCSMRT{2}{|c|}{4}{AP-GeM Aggregate}  &      M.C. &      34.38\%   &      16.75\%   &      49.50\%   &      39.00\%   &  \B{ 50.88\% }  \\
        \ocblbT                                 &      Mean &      12.40m    &       6.68m    &       3.80m    &       6.97m    &  \B{  3.30m  }  \\
        \ocblbT                                 &    Median &       4.71m    &       2.12m    &       0.50m    &       0.98m    &  \B{  0.48m  }  \\
        \ocblbT                                 &   Maximum &      50.05m    &      52.64m    &      50.02m    &  \B{ 50.01m  } &      50.02m     \\
        \hline
        \ocMCSMRT{2}{|c|}{4}{NetVLAD Aggregate} &      M.C. &      31.00\%   &      24.50\%   &      39.38\%   &      39.38\%   &  \B{ 47.12\% }  \\
        \ocblbT                                 &      Mean &      13.48m    &       5.30m    &      10.82m    &       9.03m    &  \B{  4.71m  }  \\
        \ocblbT                                 &    Median &       4.82m    &       1.27m    &       0.94m    &       0.88m    &  \B{  0.56m  }  \\
        \ocblbT                                 &   Maximum &      50.04m    &      50.04m    &      50.04m    &  \B{ 50.02m  } &  \B{ 50.02m  }  \\
        \hline
        \ocMCSMRT{2}{|c|}{4}{SALAD Aggregate}   &      M.C. &      59.00\%   &      31.62\%   &  \I{ 69.62\% } &      60.00\%   &  \B{ 67.38\% }  \\
        \ocblbT                                 &      Mean &       3.58m    &      13.17m    &       2.00m    &       1.33m    &  \B{  1.14m  }  \\
        \ocblbT                                 &    Median &       0.35m    &       1.18m    &  \I{  0.29m  } &       0.39m    &  \B{  0.30m  }  \\
        \ocblbT                                 &   Maximum &      50.04m    &      83.72m    &      50.04m    &  \B{ 50.02m  } &  \B{ 50.02m  }  \\
        \hline
    \end{tabular}
    \label{table:exp1_summary_new}
    \vspace*{-\baselineskip}
\end{table}

        Figure \ref{fig:exp1_loc_error} shows the variation in goal error for all missions where the robot reported it had arrived at the goal location; Table \ref{table:exp1_summary_new} details the mean, median and maximum goal errors. Overall for AP-GeM, the proposed MLP verification system reduces the mean goal error by $\approx73\%$ from the baseline system in completed missions, from 12.51m to 3.42m; compared to $N_P$, $N_R$, and the SVM, which see $\approx46\%$, $\approx69\%$ and $\approx43\%$ reductions respectively. This trend holds for the NetVLAD and SALAD techniques, which see $\approx65\%$ and $\approx68\%$ reductions in mean goal error from the baseline using our proposed system. Furthermore, our MLP network outperforms both naive thresholds, indicating that it has learnt statistical feature input weights that more accurately correlate with VPR integrity than simply using match distances. Figure \ref{fig:exp1_loc_error} demonstrates that the verification of VPR is especially beneficial for navigation decisions in adverse conditions, with all VPR feature types showing a significant drop in median error along the logarithmically scaled axis.
    
    \subsection{Experiment 2 Results}\label{para:exp2_results}
\begin{figure}[!ht]
            \centering
            \includegraphics[width=0.9\linewidth]{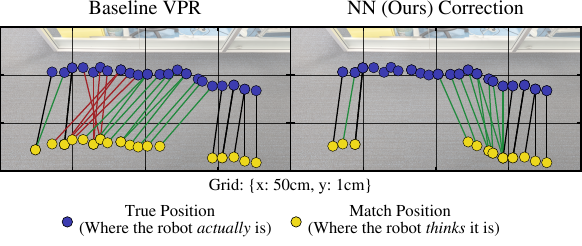}
            \caption{E.g. of the two improvement modes when utilizing the HoQ method, between baseline VPR (left) and our proposed system (right) for AP-GeM in Office Adverse. \textbf{Note:} For readability, the y-scale has been magnified. Black, red, and green correspondences denote unchanged matches, rejected matches, and corrected matches, respectively, by our proposed verification system.}
            \label{fig:exp2_along_path}
            \vspace*{-\baselineskip}
        \end{figure}

\begin{figure}[!ht]
            \centering
            \includegraphics[width=0.9\linewidth]{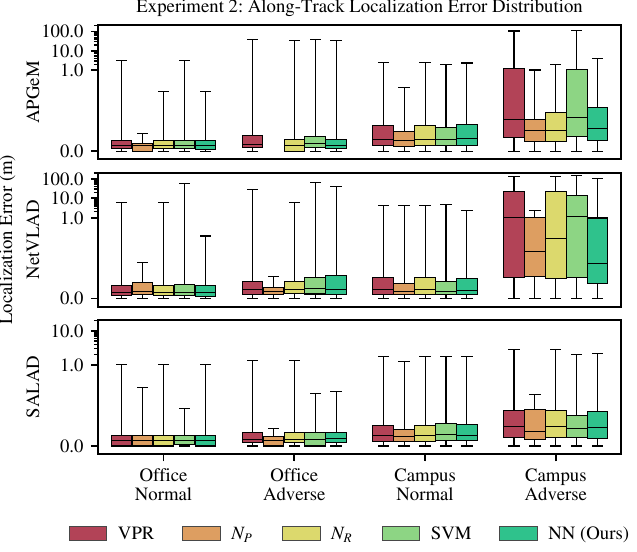}
            \caption{Localization error for all techniques, per: environment and condition. For readability, a logarithmic scale applies for high errors.}
            \label{fig:exp2_key_results_new}
        \end{figure}

        In this experiment, two modes of performance improvement are possible for a system with verification: 1) a system can choose not to localize when no verified matches are present in the recent history, or 2) a system may compute an improved localization estimate when the best \textit{verified} match is closer to the true best match. Supplementary to the forthcoming discussion, we provide real examples of these modes in Figure \ref{fig:exp2_along_path}. We provide a results summary in Figure \ref{fig:exp2_key_results_new}, with key measurements summarized in Table \ref{table:exp2table_new}.
        
        In Table \ref{table:exp2table_new}, for AP-GeM's aggregated results, we see an improvement over the baseline system in: the mean localization error from 2.29m to 0.32m ($\approx86\%$ reduction), the median localization error from 0.13m to 0.12m ($\approx8\%$ reduction), and the maximum localization error from 104.25m to 33.44m ($\approx68\%$ reduction). These improvements demonstrate our proposed system's ability to reject VPR failures with a range of severity. This trend continues for NetVLAD, where we see $\approx67\%$, $\approx14\%$, and $\approx28\%$ reductions in the aggregate mean, median, and maximum localization errors. For SALAD, the addition of our MLP verification system does not change the mean and median localization errors from the baseline system, however it does reduce the maximum error by $\approx24\%$. Given the mean and median system performance is far below the lowest training tolerance across the environments (0.5m), this is a reasonable result. We also see our proposed system improves localization error more effectively than the SVM, despite the advantage of training the SVM per environment.
        
        \begin{table}[!ht]
    \caption{Experiment 2 along-track localization error in meters. Best performance for each feature type is indicated by bold, ignoring naive thresholds. Naive thresholds are italicized if best performance.}
    \centering
    \scriptsize
    \setlength\tabcolsep{1.2mm}
    \begin{tabular}{lcL{11mm}|M{9mm}|M{9mm}M{9mm}|M{9mm}M{9mm}|}
        \cline{4-8}
        & & & \B{Baseline} & \multicolumn{4}{c|}{\B{Filtering Technique}} \\
        & & & \B{VPR}      & \boldmath{$N_P$} & \boldmath{$N_R$} & \B{SVM} & \B{Ours} \\
        \hline
        \ocMCSMRT{2}{|c|}{3}{AP-GeM Aggregate}  &      Mean &       2.29m    &  \I{  0.17m  } &       0.25m    &       2.08m    &  \B{  0.32m  }  \\
        \ocblbT                                 &    Median &       0.13m    &  \I{  0.12m  } &  \I{  0.12m  } &       0.13m    &  \B{  0.12m  }  \\
        \ocblbT                                 &   Maximum &     104.25m    &  \I{  1.04m  } &      35.21m    &     110.92m    &  \B{ 33.44m  }  \\
        \hline
        \ocMCSMRT{2}{|c|}{3}{NetVLAD Aggregate} &      Mean &       3.51m    &  \I{  0.19m  } &       2.89m    &       3.40m    &  \B{  1.16m  }  \\
        \ocblbT                                 &    Median &       0.14m    &  \I{  0.09m  } &       0.13m    &       0.13m    &  \B{  0.12m  }  \\
        \ocblbT                                 &   Maximum &     137.66m    &  \I{  4.02m  } &     137.22m    &     147.59m    &  \B{ 99.19m  }  \\
        \hline
        \ocMCSMRT{2}{|c|}{3}{SALAD Aggregate}   &      Mean &  \B{  0.19m  } &  \I{  0.14m  } &       0.19m    &       0.20m    &  \B{  0.19m  }  \\
        \ocblbT                                 &    Median &  \B{  0.12m  } &  \I{  0.10m  } &       0.12m    &  \B{  0.12m  } &  \B{  0.12m  }  \\
        \ocblbT                                 &   Maximum &       2.94m    &  \I{  1.25m  } &       2.94m    &  \B{  2.06m  } &       2.24m     \\
        \hline
    \end{tabular}
    \label{table:exp2table_new}
    \vspace*{-\baselineskip}
\end{table}
        
        \begin{table}[!ht]
    \caption{Experiment 2 system precision and recall. Best performance for each feature type is indicated by bold, ignoring naive thresholds. Naive thresholds are italicized if best performance.}
    \centering
    \scriptsize
    \setlength\tabcolsep{1.2mm}
    \begin{tabular}{lcL{11mm}|M{9mm}|M{9mm}M{9mm}|M{9mm}M{9mm}|}
        \cline{4-8}
        & & & \B{Baseline} & \multicolumn{4}{c|}{\B{Filtering Technique}} \\
        & & & \B{VPR}      & \boldmath{$N_P$} & \boldmath{$N_R$} & \B{SVM} & \B{Ours} \\
        \hline
        \ocMCSMRT{2}{|c|}{2}{AP-GeM Aggregate}  & Precision &      97.09\%   &  \I{100.00\% } &      99.95\%   &      97.05\%   &  \B{ 99.92\% }  \\
        \ocblbT                                 &    Recall &  \B{100.00\% } &      25.39\%   &      88.51\%   &      94.21\%   &      89.61\%    \\
        \hline
        \ocMCSMRT{2}{|c|}{2}{NetVLAD Aggregate} & Precision &      94.07\%   &  \I{ 99.72\% } &      95.14\%   &      94.26\%   &  \B{ 98.29\% }  \\
        \ocblbT                                 &    Recall &  \B{100.00\% } &      47.48\%   &      93.90\%   &      97.22\%   &      91.30\%    \\
        \hline
        \ocMCSMRT{2}{|c|}{2}{SALAD Aggregate}   & Precision &  \B{100.00\% } &  \I{100.00\% } &  \I{100.00\% } &  \B{100.00\% } &  \B{100.00\% }  \\
        \ocblbT                                 &    Recall &  \B{100.00\% } &      40.42\%   &      99.90\%   &      95.75\%   &      99.59\%    \\
        \hline
    \end{tabular}
    \label{table:exp2table2_new}
    \vspace*{-\baselineskip}
\end{table}

        When including the naive thresholds in the analysis, $N_P$ initially appears to outperform our proposed system across all metrics in Table \ref{table:exp2table_new}. However, by examining Table \ref{table:exp2table2_new}, we see our proposed system consistently achieves the desired balance of performance improvement with data retention (recall), whereas $N_P$ discards between ${\approx}75\%$ to ${\approx}53\%$ of the opportunities to localize even with $1.5m$ of match history. Especially considering the threshold must be selected ahead of deployment, there is no guarantee for consistent rejection of out-of-tolerance matches throughout a data set; their large variability is unsuitable for safe navigation systems. Additionally, identifying a performance point for real deployment creates another layer of complexity. An ideal verification system must be adaptive to accurately reject and retain data. As is evident in the metrics, our proposed system succeeds in removing out-of-tolerance matches regardless of match distance, surpassing all comparisons for precision whilst balancing the retention of matches as shown in the recall scores.

    \vspace{-\baselineskip}
    \subsection{Example Study: Experiment 2 Training Generalizability}\label{para:exp2_generalizability}
    
    To investigate the sensitivity of our proposed approach to training data and, therefore, the generalizability, we repeated Experiment 2 using MLP integrity monitors only trained on either the Office \textit{or} Campus training sets, rather than the Fused training set results presented earlier in the paper. Table~\ref{table:exp2_training} summarizes these results, using the mean precision and recall of each VPR technique, aggregated by the testing environment. Both the Office-only and Campus-only trained integrity monitors maintain very high precision at the cost of recall: minor for Campus-only, and significant for Office-only. Table~\ref{table:exp2_training} shows using the Fused training set retained high precision ($>99\%$) whilst sacrificing less recall. Although the Campus-only monitor generalizes better to the unseen environment (Office), likely resulting from a larger training set and training on real images, it is noteworthy that purely simulation data for Office-only achieves high precision. With a simulation dataset more representative of both environments, competitive results may be feasible.

    \begin{table}%
    \caption{Experiment 2 mean precision and recall across all VPR techniques using varied training schemes.}
    \centering
    \scriptsize
    \setlength\tabcolsep{1.2mm}
    \begin{tabular}{lcL{11mm}|M{9mm}|M{9mm}M{9mm}M{9mm}|}
        \cline{4-7}
        & & & \B{Baseline} & \multicolumn{3}{c|}{\B{NN (Ours) Training Set}} \\
        & & & \B{VPR}      & \B{Office} & \B{Campus} & \B{Fused} \\
        \hline
        \ocMCSMRT{2}{|c|}{2}{Office Aggregate}  & Precision &      99.34\%   & 100\% &      99.87\%   &      99.82\%  \\
        \ocblbT                                 &    Recall &  100\% &      57.91\%   &      94.20\%   &      97.11\%\\
        \hline
        \ocMCSMRT{2}{|c|}{2}{Campus Aggregate}  & Precision &      95.27\%   & 98.64\% &      99.36\%   &      99.09\%  \\
        \ocblbT                                 &    Recall &  100\%  &      67.14\%   &      88.91\%   &      90.71\%\\
        \hline
    \end{tabular}
    \label{table:exp2_training}
    \vspace*{-1.5\baselineskip}
\end{table}

    \vspace{-\baselineskip}
    \subsection{Stand Alone VPR Performance}
    Table~\ref{table:new_pr} provides precision, recall@1, and accuracy values, as calculated in~\cite{Schubert23_VPR_Tutorial}, for VPR performance using NetVLAD, aggregated across the QCR datasets, and for two completely unseen datasets: Nordland~\cite{Niko13} (winter reference and summer query sets) and Oxford RobotCar~\cite{RobotCarDatasetIJRR} (sunny reference and overcast query sets). Nordland queries are sampled 1:10 with reference set from~\cite{Niko13}, Oxford RobotCar sets are sampled every 1m. Here, we use the same models for our single-query method and the SVM from the above experiments, trained on the QCR datasets. The baseline performance is equal in all metrics because it accepts all VPR matches. Values surpassing the baseline are in bold, otherwise the best verification method is underlined. As seen, our verification system, in general, has the highest precision and accuracy, with varied reductions in recall. This indicates the model has some generalizability to data from a totally different distribution to the training data.

\begin{table}[!ht]
    \caption{Precision, recall@1, and accuracy for VPR performance across datasets. The baseline comparison `accepts' all VPR position estimates. The models trained for the above experiments on the QCR training sets are used across all datasets. We compute precision and recall@1 for VPR as in~\cite{Schubert23_VPR_Tutorial}.}
    \centering
    \scriptsize
    \setlength\tabcolsep{1.2mm}
    \begin{tabular}{|L{7mm}|M{6mm}|M{6mm}M{6mm}|M{6mm}|M{6mm}M{6mm}|M{6mm}|M{6mm}M{6mm}|}
        \hline
        \textbf{Net-} & \multicolumn{3}{c|}{\B{QCR Aggregate}}  & \multicolumn{3}{c|}{\B{Nordland}~\cite{Niko13}} & \multicolumn{3}{c|}{\B{RobotCar}~\cite{RobotCarDatasetIJRR}} \\
        \cline{2-10}
        \textbf{VLAD} & \B{VPR}  & \B{SVM} & \B{Ours} & \B{VPR} & \B{SVM} & \B{Ours}& \B{VPR} & \B{SVM} & \B{Ours} \\
        \hline
        \multicolumn{1}{|c|}{Prec.}    &  75.4\%             & 77.8\%   & \textbf{87.2\%}     &   8.6\%   & 7.0\%              &  \textbf{13.6\%}   &      77.0\%  & 71.4\%    &     \textbf{83.1\%}  \\
        \multicolumn{1}{|c|}{Rec@1}     &  75.4\%              & 24.3\%   & \underline{49.7\%}  &   8.6\%   & \underline{6.8\%} &    4.02\%          &      77.0\%   & 53.4\%    &     \underline{67.9\%}  \\
        \multicolumn{1}{|c|}{Acc.}     &  75.4\%    & 41.9\%   & \underline{69.0\%}              &   8.6\%   & 8.3\%              &  \textbf{69.8\%}   &      77.0\%    & 54.9\%    &    \textbf{77.1\%}  \\
        \hline
    \end{tabular}
    \label{table:new_pr}
    \vspace*{-\baselineskip}
\end{table}
    \section{Conclusion}
    
    Providing metrics that convey confidence or uncertainty will continue to be a key requirement for trustworthy systems across robotics. This letter presents a real-time integrity-based verification method that can reject VPR errors which may lead to unsafe navigation decisions. We demonstrate two implementations: (a) a single-query rejection method in the context of a robot navigating to a goal zone, and (b) a history-of-queries method that takes a best, verified match and uses an odometer to extrapolate forwards to a position estimate.
    
    We found that our proposed system offers a performance improvement, with some noteworthy results. In Experiment 1, we see the aggregate number of missions successfully completed increase from the baseline's \expOneAggMCVPR~ to \expOneAggMCNN~ for our proposed system, with aggregate mean goal error reductions from baseline of \mbox{$\approx\!73\%$}, \mbox{$\approx\!65\%$}, and \mbox{$\approx68\%$} for the AP-GeM, NetVLAD, and SALAD techniques respectively. In Experiment 2, our proposed system sees reductions in the aggregate mean localization error reductions from baseline of \mbox{$\approx\!86\%$} and \mbox{$\approx\!67\%$} for AP-GeM and NetVLAD, and a reduction in maximum localization error by \mbox{$\approx\!24\%$} for SALAD which was already operating within tolerance for mean localization error. We additionally see a minimum of \mbox{$\approx\!98.3\%$} precision for our proposed system across all technique aggregates, up from the baseline's value at \mbox{$\approx\!94.1\%$}.

    Although our proposed system cannot yet quantitatively estimate error (either lateral or longitudinal), there already exists scenarios where the presented capabilities can be useful. They may provide redundancy, correcting signals, or trigger events such as loop closure in SLAM systems. With our proposed system serving as a demonstrator, further development could see improved heuristics and feature selection for MLP training, training data curation across larger sets, error-binned classes, a weighting scheme, or advanced neural network implementations. For the HoQ method, applying weights to history entries may improve performance, which could be either hand-crafted from heuristics or learnt. An example would be promoting recent matches over older matches, which may be especially useful for applications with long histories.
    
    \vspace{-0.5\baselineskip}

    \bibliographystyle{IEEEtran}
    \bibliography{IEEEabrv,ref.bib}

\begin{thebibliography}{10}
\providecommand{\url}[1]{#1}
\csname url@samestyle\endcsname
\providecommand{\newblock}{\relax}
\providecommand{\bibinfo}[2]{#2}
\providecommand{\BIBentrySTDinterwordspacing}{\spaceskip=0pt\relax}
\providecommand{\BIBentryALTinterwordstretchfactor}{4}
\providecommand{\BIBentryALTinterwordspacing}{\spaceskip=\fontdimen2\font plus
\BIBentryALTinterwordstretchfactor\fontdimen3\font minus
  \fontdimen4\font\relax}
\providecommand{\BIBforeignlanguage}[2]{{%
\expandafter\ifx\csname l@#1\endcsname\relax
\typeout{** WARNING: IEEEtran.bst: No hyphenation pattern has been}%
\typeout{** loaded for the language `#1'. Using the pattern for}%
\typeout{** the default language instead.}%
\else
\language=\csname l@#1\endcsname
\fi
#2}}
\providecommand{\BIBdecl}{\relax}
\BIBdecl

\bibitem{Jacobson21}
A.~Jacobson, F.~Zeng, D.~Smith, N.~Boswell, T.~Peynot, and M.~Milford, ``What
  localizes beneath: A metric multisensor localization and mapping system for
  autonomous underground mining vehicles,'' \emph{J. Field Robot.}, 2021.

\bibitem{Lowry2016}
S.~Lowry and et~al., ``Visual place recognition: A survey,'' \emph{IEEE Trans.
  Robot.}, Feb. 2016.

\bibitem{Schubert23_VPR_Tutorial}
S.~Schubert, P.~Neubert, S.~Garg, M.~Milford, and T.~Fischer, ``Visual place
  recognition: A tutorial,'' \emph{IEEE Robot. Automat. Mag.}, 2023.

\bibitem{Masone2021}
C.~Masone and B.~Caputo, ``A survey on deep visual place recognition,''
  \emph{IEEE Access}, 2021.

\bibitem{NetVLAD18}
R.~Arandjelović, P.~Gronat, A.~Torii, T.~Pajdla, and J.~Sivic, ``Netvlad: Cnn
  architecture for weakly supervised place recognition,'' \emph{IEEE Trans.
  Pattern Anal. Mach. Intell.}, 2018.

\bibitem{Mereu2022}
R.~Mereu, G.~Trivigno, G.~Berton, C.~Masone, and B.~Caputo, ``Learning
  sequential descriptors for sequence-based visual place recognition,'' Jul
  2022, arXiv:2207.03868.

\bibitem{Wang22_CVPR}
R.~Wang, Y.~Shen, W.~Zuo, S.~Zhou, and N.~Zheng, ``Transvpr: Transformer-based
  place recognition with multi-level attention aggregation,'' in \emph{IEEE/CVF
  Conf. Comput. Vis. Pattern Recognit.}, 2022.

\bibitem{PatchNetVLAD}
S.~Hausler, S.~Garg, M.~Xu, M.~Milford, and T.~Fischer, ``Patch-netvlad:
  Multi-scale fusion of locally-global descriptors for place recognition,'' in
  \emph{IEEE/CVF Conf. Comput. Vis. Pattern Recognit.}, 2021.

\bibitem{CosPlace}
G.~Berton, C.~Masone, and B.~Caputo, ``Rethinking visual geo-localization for
  large-scale applications,'' in \emph{IEEE/CVF Conf. Comput. Vis. Pattern
  Recognit.}, 2022.

\bibitem{EigenPlaces23}
G.~Berton, G.~Trivigno, B.~Caputo, and C.~Masone, ``Eigenplaces: Training
  viewpoint robust models for visual place recognition,'' in \emph{IEEE/CVF
  Int. Conf. Comput. Vis.}, 2023.

\bibitem{Lsplacerecgss}
M.~Leyva-Vallina, N.~Strisciuglio, and N.~Petkov, ``Data-efficient large scale
  place recognition with graded similarity supervision,'' in \emph{IEEE/CVF
  Conf. Comput. Vis. Pattern Recognit.}, 2023.

\bibitem{MixVPR23}
A.~Ali-Bey, B.~Chaib-Draa, and P.~Giguére, ``Mixvpr: Feature mixing for visual
  place recognition,'' in \emph{IEEE/CVF Winter Conf. Appl. Comput. Vis.},
  2023.

\bibitem{optimalsalad}
S.~Izquierdo and J.~Civera, ``Optimal transport aggregation for visual place
  recognition,'' in \emph{Proc. IEEE/CVF Conf. Comput. Vis. Pattern Recognit.},
  2023.

\bibitem{anyloc}
N.~Keetha and et~al., ``Anyloc: Towards universal visual place recognition,''
  \emph{IEEE Robot. Autom. Lett.}, 2023.

\bibitem{2021Schubert}
S.~Schubert and P.~Neubert, ``What makes visual place recognition easy or
  hard?'' Jun. 2021, arXiv:2106.12671.

\bibitem{zaffar2024estimation}
M.~Zaffar, L.~Nan, and J.~F. Kooij, ``On the estimation of image-matching
  uncertainty in visual place recognition,'' in \emph{Proc. IEEE/CVF Conf.
  Comput. Vis. Pattern Recognit.}, 2024.

\bibitem{Zhu22}
C.~Zhu, M.~Meurer, and C.~Günther, ``Integrity of visual navigation —
  developments, challenges, and prospects,'' \emph{Navig.}, 6 2022.

\bibitem{trivigno2023divide}
G.~Trivigno, G.~Berton, J.~Aragon, B.~Caputo, and C.~Masone,
  ``Divide\&classify: Fine-grained classification for city-wide visual
  geo-localization,'' in \emph{Proc. IEEE/CVF Int. Conf. Comput. Vis.}, 2023.

\bibitem{Stenborg20}
E.~Stenborg, T.~Sattler, and L.~Hammarstrand, ``Using image sequences for
  long-term visual localization,'' in \emph{2020 Int. Conf. 3D Vis.}, 2020.

\bibitem{Yin19}
P.~Yin and et~al., ``Mrs-vpr: a multi-resolution sampling based global visual
  place recognition method,'' in \emph{Int. Conf. Robot. Automat.}, 2019.

\bibitem{Tsintotas2021}
K.~A. Tsintotas, L.~Bampis, and A.~Gasteratos, ``Tracking-doseqslam: A dynamic
  sequence-based visual place recognition paradigm,'' \emph{IET Comput. Vis.},
  6 2021.

\bibitem{Xu21}
M.~Xu, N.~Sunderhauf, and M.~Milford, ``Probabilistic visual place recognition
  for hierarchical localization,'' \emph{IEEE Robot. Automat. Lett.}, 2021.

\bibitem{Warburg21}
F.~Warburg, M.~Jørgensen, J.~Civera, and S.~Hauberg, ``Bayesian triplet loss:
  Uncertainty quantification in image retrieval,'' in \emph{IEEE/CVF Int. Conf.
  Comput. Vis.}, 2021.

\bibitem{Warburg23}
F.~Warburg, M.~Miani, S.~Brack, and S.~Hauberg, ``Bayesian metric learning for
  uncertainty quantification in image retrieval,'' 2023.

\bibitem{Li15}
J.~Li, R.~M. Eustice, and M.~Johnson-Roberson, ``Underwater robot visual place
  recognition in the presence of dramatic appearance change,'' in \emph{OCEANS
  2015 - MTS/IEEE Washington}, 2015.

\bibitem{particlefilters18}
P.~Karkus, D.~Hsu, and W.~S. Lee, ``Particle filter networks with application
  to visual localization,'' in \emph{Conf. Robot Learn.}, 2018.

\bibitem{hausler2021unsupervised}
S.~Hausler, T.~Fischer, and M.~Milford, ``Unsupervised complementary-aware
  multi-process fusion for visual place recognition,'' 2021, arXiv:2112.04701.

\bibitem{Malone23}
C.~Malone, S.~Hausler, T.~Fischer, and M.~Milford, ``Boosting performance of a
  baseline visual place recognition technique by predicting the maximally
  complementary technique,'' in \emph{IEEE Int. Conf. Robot. Automat.}, 2023.

\bibitem{Hage21}
J.~A. Hage, P.~Xu, P.~Bonnifait, and J.~Ibañez-Guzmán, ``Localization
  integrity for intelligent vehicles through fault detection and position error
  characterization,'' \emph{IEEE Trans. Intell. Transp. Systems}, 2022.

\bibitem{Li2019}
C.~Li and S.~L. Waslander, ``Visual measurement integrity monitoring for uav
  localization,'' in \emph{IEEE Int. Symp. Saf., Secur., Rescue Robot.}, 2019.

\bibitem{Wang20}
S.~Wang, X.~Zhan, Y.~Fu, and Y.~Zhai, ``\BIBforeignlanguage{eng}{Feature-based
  visual navigation integrity monitoring for urban autonomous platforms},''
  \emph{\BIBforeignlanguage{eng}{Aerosp. systems}}, 2020.

\bibitem{Fu2020}
Y.~Fu and et~al., ``Visual odometry errors and fault distinction for integrity
  monitoring,'' \emph{Aerosp. Systems}, 2020.

\bibitem{Fu22}
Y.~Fu, S.~Wang, Y.~Zhai, X.~Zhan, and X.~Zhang, ``Measurement error detection
  for stereo visual odometry integrity,'' \emph{Navig.}, 2022.

\bibitem{Gabela20}
J.~Gabela, I.~Majic, A.~Kealy, M.~Hedley, and S.~Li, ``Robust vehicle
  localization and integrity monitoring based on spatial feature constrained
  pf,'' in \emph{IEEE/ION Position, Location Navig. Symp.}, 2020.

\bibitem{Calhoun16}
S.~M. Calhoun and J.~Raquet, ``Integrity determination for a vision based
  precision relative navigation system,'' in \emph{IEEE/ION Position, Location
  Navig. Symp.}, 2016.

\bibitem{Arana2020}
G.~D. Arana, O.~A. Hafez, M.~Joerger, and M.~Spenko, ``Localization safety
  validation for autonomous robots,'' \emph{IEEE/RSJ Int. Conf. Intell. Robots
  Systems}, 2020.

\bibitem{Balakrishnan2020}
A.~Balakrishnan, S.~R. Florez, and R.~Reynaud, ``Integrity monitoring of
  multimodal perception system for vehicle localization,'' \emph{Sensors
  (Switzerland)}, 8 2020.

\bibitem{AlHage19}
J.~Al~Hage, P.~Xu, and P.~Bonnifait, ``\BIBforeignlanguage{eng}{High integrity
  localization with multi-lane camera measurements},'' in
  \emph{\BIBforeignlanguage{eng}{IEEE Intell. Vehicles Symp.}}, 2019.

\bibitem{Bhamidipati21}
S.~Bhamidipati and G.~X. Gao, ``Robust gps-vision localization via
  integrity-driven landmark attention,'' Jan 2021, arXiv:2101.04836.

\bibitem{Carson22}
H.~Carson, J.~J. Ford, and M.~Milford, ``Predicting to improve: Integrity
  measures for assessing visual localization performance,'' \emph{IEEE Robot.
  Automat. Lett.}, 2022.

\bibitem{Revaud2019APGEM}
J.~Revaud, J.~Almaz{\'a}n, R.~S. Rezende, and C.~R.~d. Souza, ``Learning with
  average precision: Training image retrieval with a listwise loss,'' in
  \emph{IEEE/CVF Int. Conf. Comput. Vis.}, 2019.

\bibitem{hdl_packages}
K.~Koide, J.~Miura, and E.~Menegatti, ``A portable three-dimensional
  lidar-based system for long-term and wide-area people behavior measurement,''
  \emph{Int. J. Adv. Robot. Systems}, 02 2019.

\bibitem{Niko13}
N.~Sünderhauf, P.~Neubert, and P.~Protzel, ``Are we there yet? challenging
  seqslam on a 3000 km journey across all four seasons,'' \emph{Int. Conf.
  Robot. Automat.}, 2013.

\bibitem{RobotCarDatasetIJRR}
W.~Maddern, G.~Pascoe, C.~Linegar, and P.~Newman, ``{1 Year, 1000km: The Oxford
  RobotCar Dataset},'' \emph{Int. J. Robot. Res.}, 2017.

\end{thebibliography}

    \vfill
\end{document}